\documentclass[]{article}
\usepackage[letterpaper]{geometry}
\usepackage{mtsummit2015}
\usepackage{times}
\usepackage{url}
\usepackage{latexsym}
\usepackage{natbib}
\usepackage{layout}
\usepackage{graphicx}
\usepackage{subcaption}

\begin{document}

\title{\bf Fluency Over Adequacy: A Pilot Study in Measuring User Trust in Imperfect MT}  

\author{\name{\bf Marianna J. Martindale} \hfill  \addr{mmartind@umd.edu}\\ 
        \addr{iSchool, University of Maryland, 
        College Park, 20740, USA}
\AND
       \name{\bf Marine Carpuat} \hfill \addr{marine@cs.umd.edu}\\
        \addr{Dept. of Computer Science, University of Maryland, 
        College Park, 20740, USA}
}        
\maketitle
\pagestyle{empty}

\begin{abstract}
Although measuring intrinsic quality has been a key factor in the advancement of Machine Translation (MT), successfully deploying MT requires considering not just intrinsic quality but also the user experience, including aspects such as trust. This work introduces a method of studying how users modulate their trust in an MT system after seeing errorful (disfluent or inadequate) output amidst good (fluent and adequate) output. We conduct a survey to determine how users respond to good translations compared to translations that are either adequate but not fluent, or fluent but not adequate. In this pilot study, users responded strongly to disfluent translations, but were, surprisingly, much less concerned with adequacy.

\end{abstract}

\section{Introduction}

Machine translation has reached a level of quality such that it can now provide fast, cheap, ``good enough'' translations for a wide variety of information sources in an increasingly diverse set of languages and dialects. These systems are becoming available for real users who have no other way to understand content in languages in which they are not proficient. While the overall quality of MT output is important, in practice the utility of MT also depends on the willingness of users to accept the technology. 

Trust is an important factor in that willingness regardless of use case. In the case of MT for communication, users may choose not to communicate if they do not trust the MT available to them. If translators do not trust the MT in their computer-assisted translation environment, they may be less efficient. When there is a need for MT for understanding, insufficient trust in MT may lead the user to simply avoid content in languages they do not understand. However lack of trust is not the only issue. Implicit trust in incorrect MT can be at least as problematic. Users may misunderstand inaccurately translated information, an automatically translated website could reflect badly on the company, and inaccurate translations used in communication could be embarrassing. However, there could be more serious repercussions if the MT is used in critical tasks. This is not merely theoretical: In October 2017, a Palestinian man was mistakenly arrested by Israeli police based on an inaccurate Facebook translation \citep{haaretz2017}.

Because trust is so important, we need to understand how users modulate their trust in MT. However trust has been a neglected element of MT evaluation to date.  To be most effective, trust evaluation must be taken into consideration not only at deployment time, but also during research and development. Understanding what factors affect user trust can help identify the most problematic error types, informing research directions. In this paper, we introduce a method of studying user trust in MT in a lab setting, based on previous work on trust in other forms of automation. We apply this approach in a pilot study comparing the effects of fluency and adequacy errors on user trust, and find that fluency appears to have a much greater effect than adequacy.

\section{Related Work}

To the best of our knowledge, there has been no prior work focused on evaluating user trust in MT. Automatically generated confidence scores have been referred to as ``trust'' \citep{Soricut:2010:TIT:1858681.1858744}, but user trust has not been evaluated. There has, however, been extensive work in human evaluation of MT quality and in user trust of other types of automation.

\subsection{Human Evaluation of Machine Translation}

Automated quality metrics such as BLEU \citep{papineni_bleu:_2002} and METEOR \citep{banerjee_meteor:_2005} have been invaluable for improving and training MT systems, and have been shown to correlate with human rankings at the system level \citep{papineni_bleu:_2002}, but human judgments remain the gold standard for competitions like the Workshop (now Conference) on Machine Translation (WMT) \citep{callison-burch_meta-_2007,callison-burch_further_2008,callison-burch_findings_2009,callison-burch_findings_2010,callison-burch_findings_2011,callison-burch_findings_2012,bojar_findings_2013,bojar_findings_2014,bojar_findings_2015,bojar_findings_2016,bojar_findings_2017}. However, these human judgments have focused almost entirely on intrinsic quality rather than the user experience and particularly trust. 

Common approaches to human evaluation of MT include ranking and direct assessment. Ranking was the official metric for WMT 2008 - 2016 \citep{callison-burch_further_2008,callison-burch_findings_2009,callison-burch_findings_2010,callison-burch_findings_2011,callison-burch_findings_2012,bojar_findings_2013,bojar_findings_2014,bojar_findings_2015,bojar_findings_2016}. In each evaluation, participants ranked five translations of individual segments (usually sentences) from different MT systems against each other based on a reference translation. While rankings indicate user preference, they do not explicitly address user trust in the system and do not directly provide insight into aspects of quality that affect user judgments of the translation. The setting is also very different from a typical user experience, in which they would see only one translation of a whole document with no reference translation rather than multiple translations of individual segments and a human translation to compare against. This difference makes it difficult to draw conclusions about the user experience.

Direct assessment of adequacy and fluency is slightly more like the typical user experience in that they only see one translation. WMT initially adopted a direct assessment approach for the 2006 and 2007 workshops \citep{koehn_manual_2006,callison-burch_meta-_2007}, but the inter-annotator agreement was low that the metric was abandoned until WMT16 \citep{bojar_findings_2016}. The change came as a result of a pilot direct assessment, conducted in parallel to the official rankings, that tested the technique from \citet{graham_continuous_2013} on a larger scale. Graham et al. had found that they could improve inter-annotator agreement by accounting for the personal preferences of the judges: They used a near-continuous scale (100 points instead of only five) and standardized each judge's score to a z-score based on the mean and standard deviation of that judge's scores. The results in the WMT16 pilot correlated so well that WMT17 used direct assessment of adequacy as the official metric \citep{bojar_findings_2017}. However, the measure is still focused on intrinsic quality and the judges still only see one segment at a time rather than a whole document. 

Although there is much less research on evaluating the user experience with MT than there is on evaluating MT quality, there have been a few user studies of MT. User studies have looked directly at how MT is used in assimilation, dissemination, and communication contexts, but have not focused on trust. For example, \citet{yang_systran_1998} looked at user reactions to the first free online MT system, AltaVista's Babelfish (powered by SYSTRAN). They analyzed feedback sent to the company during the first five months the system was available. They found the use was largely assimilation, but there were also examples of using the system for dissemination (e.g., websites using the service to provide translations of their own site), and communication (e.g., interacting with relatives or employees who don't speak English). In addition to providing insights into the applications of the system, the feedback provides positive or negative reactions. The study does not address trust or acceptance of the system, and it would be difficult to determine from this feedback since those who chose to contact the company may not be a representative sample of the pool of potential users.

\citet{hara_effect_2015} conducted a user study of MT for communication. They used spoken
language translation, which adds speech-to-text and text-to-speech technology  to the MT. Participants were not asked to directly rate fluency and adequacy of the MT, but in the feedback some mentioned grammatical and word order errors as well as problems with idiomatic expressions. While they discussed the effects these errors had on their ability to communicate, they did not address trust in the system.

To the best of our knowledge, the only published research specifically mentioning user trust in MT is \citet{karamanis_translation_2011}, who conducted a qualitative study of translation practices at a language service provider, including the use of MT. They observed that translators trusted translations from team members more than translations from remote freelancers and that they trusted MT and information gleaned from web searches even less. This study was focused on the actual work practices of the translators and the high-level interactions with MT and other translators, so it does not investigate why they distrust MT or whether they would trust one MT system over another. This dissemination setting is also different from an online MT for assimilation setting not only in the purpose but in the nature of the users. While translators may use MT to translate more efficiently, for the most part they are able to translate with their own knowledge and other resources. Those who do not know the source language are dependent on the MT to understand the text. These differences may affect their trust in the system.

\subsection{User Trust in Automation} \label{trustinauto}

Although trust has been largely neglected in the context of MT, it has been  studied in the context of other types of automation. It has been measured in a number of ways from simply asking the user \citep{yang_how_2016} to measuring the user's expectations of system performance and actions taken \citep{de_vries_effects_2003}. 

In \citet{de_vries_effects_2003}, they used a route-planning scenario to see what effect errors would have on users\rq\, trust of the system. They define trust as a user's expectations of how a system will perform in a situation of uncertainty with some risk associated. Participants were asked to perform 26 route-planning tasks. For each task, they were asked to stake a bet on how good the route would be and they were told that the credits they earned across the whole task would affect their compensation. Ten of these trials were manual, ten automated, and in the final six the participant was allowed to choose manual or automatic. Trust was measured based on the user's stated trust in the system on a five-point scale, the number of times they chose the automated system in the last six trials, and the amount they wagered on each trial. They found that all things being equal, users preferred manual mode, but when the automated system made more errors, they trusted the automated system even less according to all measures.

Scenarios based on perception have also been used to explore user trust in automated systems, e.g. \citet{dzindolet_role_2003} and \citet{yang_how_2016}. The latter is particularly relevant to this work. They  looked at how negative experiences can affect trust in automation for a face recognition task. Participants were shown a set of photos of target faces to remember. They were later shown a target photo and a distractor and were asked to identify the target photo. They were evaluated based on their performance on the task. They were also provided with a suggested answer which they could choose to accept or reject. Users self-reported their trust in the system using a six-point scale. If trust were a purely statistical process, an incorrect suggestion by the system should have the same effect on trust whether or not the user chose to accept the system's suggestion. However, they observed that when the participant accepted the system's suggestion and then learned that the answer was wrong, their trust in the system dropped much more than when they rejected the system's incorrect suggestion.

\section{Assessing Trust in MT}

Like the examples in Section \ref{trustinauto}, MT is a kind of automation. It performs a typically human task (translation) and the results can be good or bad in terms of both fluency and adequacy. User reactions to an MT system may change over time as the user is exposed to more translations and the resulting change in trust of those interactions may not correspond with a simple statistical measure like intrinsic quality. In the assimilation setting, users will not typically know the adequacy of a translation but can readily judge its fluency. This means fluency may have an immediate effect on trust while adequacy will only have an effect if the user subsequently learns that the translation was wrong.

We can think of a user's trust in a specific machine translation like the wagers in \citet{de_vries_effects_2003} or the choice to agree or disagree with the system prediction in \citet{yang_how_2016}. As users see translations that are disfluent or inadequate, we expect that their trust in the system will decrease. A situation where the user trusts a translation based on fluency but then subsequently learns that it was not adequate could be compared to the case in \citet{yang_how_2016} where users accept the system prediction and then learn that it was incorrect. If the findings in \citeauthor{yang_how_2016} also apply to the machine translation scenario, we would expect that user trust in the MT system will go down further when they see this type of bad translation compared to translations that are
merely disfluent. 

Because these fluent-but-not-adequate translations may lead the user to an incorrect conclusion, we will refer to them as \textit{misleading}. Although misleading translations are rare--only 434 out of 43,911 segments (0.99\%) in system output submissions for WMT16 \citep{bojar_findings_2016}, the potential for mistakes in understanding and degradation of user trust makes the translations particularly problematic when they occur. This is especially important as the MT community moves to neural machine translation (NMT), as NMT has been shown to consistently provide more fluent translations than previous MT paradigms \citep{bentivogli_neural_2016,toral_multifaceted_2017,koehn_six_2017}, but is also prone to producing output that is fluent but unrelated to the input when there is out of domain or insufficient training data \citep{koehn_six_2017}.

We propose a survey-based approach to see how users react to misleading translations as compared to their reactions to disfluent translations. Our pilot study seeks to evaluate the following hypotheses:
\begin{itemize}
\item[]\textbf{H1:} Exposure to good translations will maintain or increase user trust in the system
\item[]\textbf{H2:} Exposure to bad MT output of either type will decrease user trust in the system 
\item[]\textbf{H3:} Misleading translations will have a bigger effect on trust than disfluent translations
\end{itemize}

For the purposes of this study, we focus on an assimilation use case, reading foreign language news. Assimilation tasks are the most practical to implement as they do not require specialized skills or interaction between participants. We chose the news domain because it is an established domain for MT research and output from state-of-the art systems is available through WMT. 

\section{Survey Methodology}

The task of \textit{studying trust in MT} differs from the task of \textit{evaluating MT quality} in several important ways. MT quality evaluations are focused on a specific system or a comparison between systems. This requires many segments each evaluated by several judges. In effect, the segments of MT output are a sample from the population of all segments that could be translated by the system. Trust could potentially be used as an additional dimension in such an evaluation. However, it is valuable to study the phenomenon of user trust in MT in order to know how to help users have an appropriate level of trust in MT output. Studying the phenomenon of trust in MT is focused in the opposite direction of system evaluations: the users are the population that is randomly sampled and the segments cannot be randomly selected but must be carefully chosen to demonstrate the specific properties being compared while mediating confounding variables. 

\paragraph{Survey Overview} For this study, we obtained Institutional Review Board approval for and conducted a survey in which participants were shown outputs from a hypothetical MT system constructed from actual MT output. In order to establish a baseline trust level and to account for biases, the participants were first asked their familiarity with machine and human translation as well as their overall opinion of machine translation. After these initial questions, they were shown a series of five machine translation outputs and asked to rate their level of trust. Presenting a series of outputs makes it possible to study change in trust over consecutive translation examples. For each translation, they first saw only the MT output and were asked their level of trust in that specific translation and their trust in the overall system based on all of the translations they had seen so far. They were then shown a human translation of the text and asked whether their judgment of the MT output had changed as well as their overall trust of the system.

\begin{figure}[t]
\centering
\includegraphics[width=0.75\textwidth]{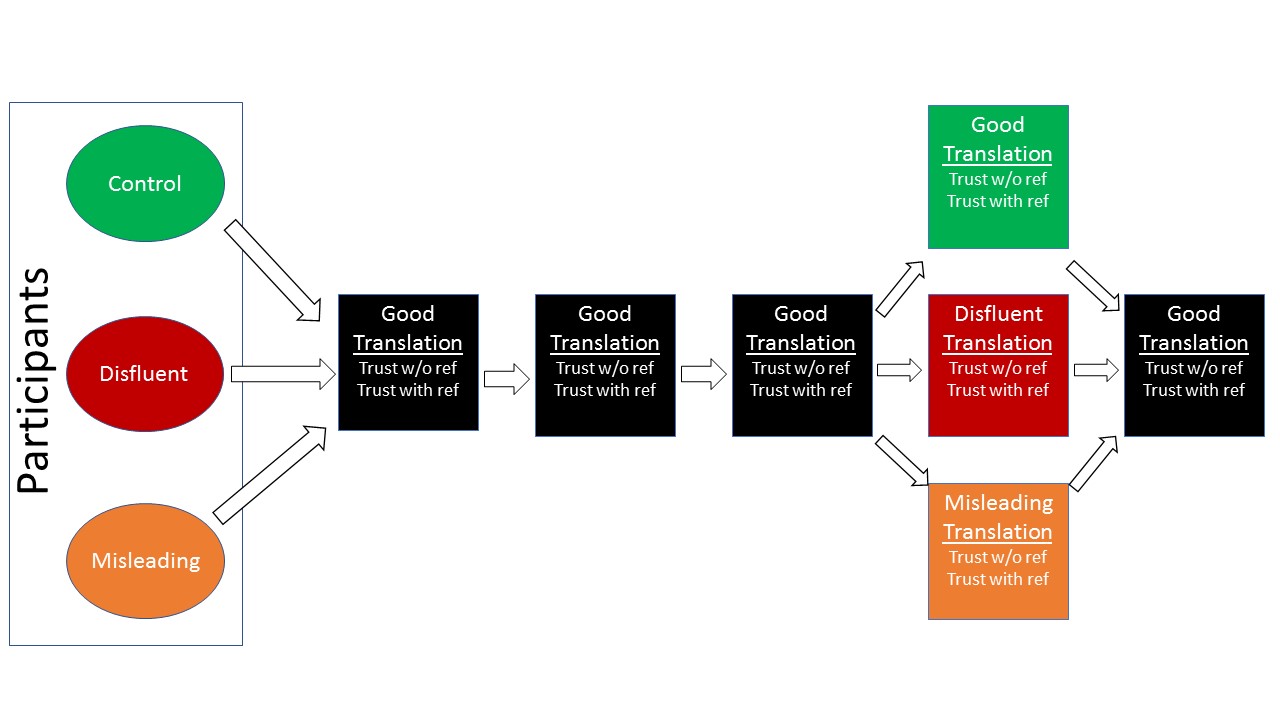}
\caption{Diagram illustrating the flow of the survey for all three groups.}
\label{fig1}
\end{figure}

\paragraph{Three Conditions} All participants saw the same three good (fluent and adequate) translations in the same order to establish a baseline trust level. To test the impact of misleading translations as opposed to good translations or obviously bad translations, the participants were split into three groups for the fourth translation: a control group which saw another good translation; a disfluent group which saw a disfluent but adequate translation; and a misleading group which saw the misleading translation. These alternate translations were translations of the same passage by different MT systems to ensure that fluency and adequacy were the only difference and not topic or complexity of the source. Finally, all participants were shown the same good translation as their fifth translation to see if changes in trust would carry over to the next translation. The flow of the survey is illustrated in Figure \ref{fig1}. 

\paragraph{Comprehension Questions for Quality Control} To ensure participants were paying attention and understood the translation examples, they were also asked to answer a simple comprehension question for each translation before seeing the human translation. This consisted of a choice between two gists of the translation with opposite meanings. For the misleading translation, the “correct” answer was what the machine translation said rather than what it should have said. This not only ensured that participants were being misled but also primed them with the misleading meaning to increase the chance that they would feel misled when they saw the human translation. An example question is shown in Figure \ref{fig2}.

\subsection{Participants}

\begin{table}[t]
\centering
\begin{tabular}{ | c | c | c | c | }
\hline
 \textbf{Group} & \textbf{No Familiarity} & \textbf{Some Familiarity} & \textbf{Very Familiar} \\ 
 \hline
 \textbf{Control} & 60.1\% & 20.7\% & 17.2\% \\  
 \hline
 \textbf{Disfluent} & 40.6\% & 34.4\% & 25.0\% \\
 \hline
  \textbf{Misleading} & 57.1\% & 17.9\% & 25.0\% \\
 \hline
  \textbf{All} & 52.8\% & 24.7\% & 22.5\% \\
 \hline
\end{tabular}
\caption{Participant familiarity with machine translation}
\label{table1}
\bigskip
\bigskip
\begin{tabular}{ | c | c | c | c | }
\hline
 \textbf{Group} & \textbf{No Familiarity} & \textbf{Some Familiarity} & \textbf{Very Familiar} \\ 
 \hline
 \textbf{Control} & 65.5\% & 31.0\% & 3.5\% \\  
 \hline
 \textbf{Disfluent} & 78.1\% & 15.6\% & 6.3\% \\
 \hline
  \textbf{Misleading} & 78.5\% & 17.9\% & 3.6\% \\
 \hline
  \textbf{All} & 74.2\% & 21.3\% & 4.5\% \\
 \hline
\end{tabular}
\caption{Participant familiarity with human translation}
\label{table2}
\bigskip
\bigskip
\begin{tabular}{ | c | c | c | c | }
\hline
 \textbf{Group} & \textbf{Negative} & \textbf{Neutral} & \textbf{Positive} \\ 
 \hline
 \textbf{Control} & 13.8\% & 51.7\% & 34.5\% \\  
 \hline
 \textbf{Disfluent} & 3.1\% & 65.6\% & 31.3\% \\
 \hline
  \textbf{Misleading} & 7.1\% & 35.7\% & 57.1\% \\
 \hline
  \textbf{All} & 7.9\% & 51.7\% & 40.4\% \\
 \hline
\end{tabular}
\caption{Participant overall perception of MT}
\label{table3}
\end{table}

To represent typical, non-specialist MT users, participants were recruited via Qualtrics from their pool of participants over 18 and located in the U.S. Participants were not asked their age, gender, or native language, as age and gender are unlikely to be relevant in understanding machine translation output and the comprehension questions would serve to judge participants\rq\,  English language ability. Any responses that appeared to be invalid (e.g., giving the same value for every question) were excluded. Of the remaining responses, only responses that correctly answered all of comprehension questions for the four good translations were included in the final analysis. From the control and misleading groups, those who incorrectly answered the comprehension question for the fourth translation were also excluded. For the disfluent group, the fourth comprehension question was ignored because the lack of fluency made the translation so difficult to understand that half of the participants who answered all other comprehension questions correctly responded incorrectly to the question. 

\begin{figure}[h]
    \centering
    \begin{minipage}{0.45\textwidth}
        \centering
		\includegraphics[width=0.99\textwidth]{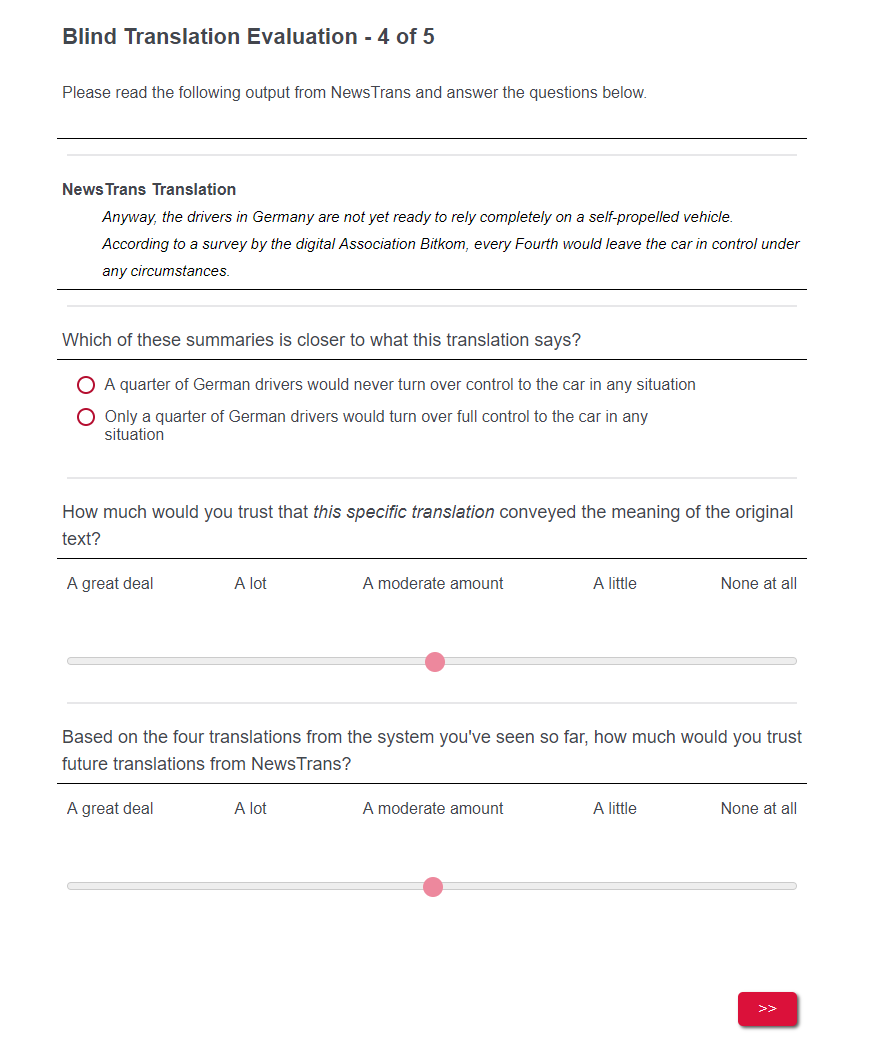}
		\caption{Example survey question}
		\label{fig2}
    \end{minipage}\hfill
    \begin{minipage}{0.45\textwidth}
        \centering
        \includegraphics[width=0.99\textwidth]{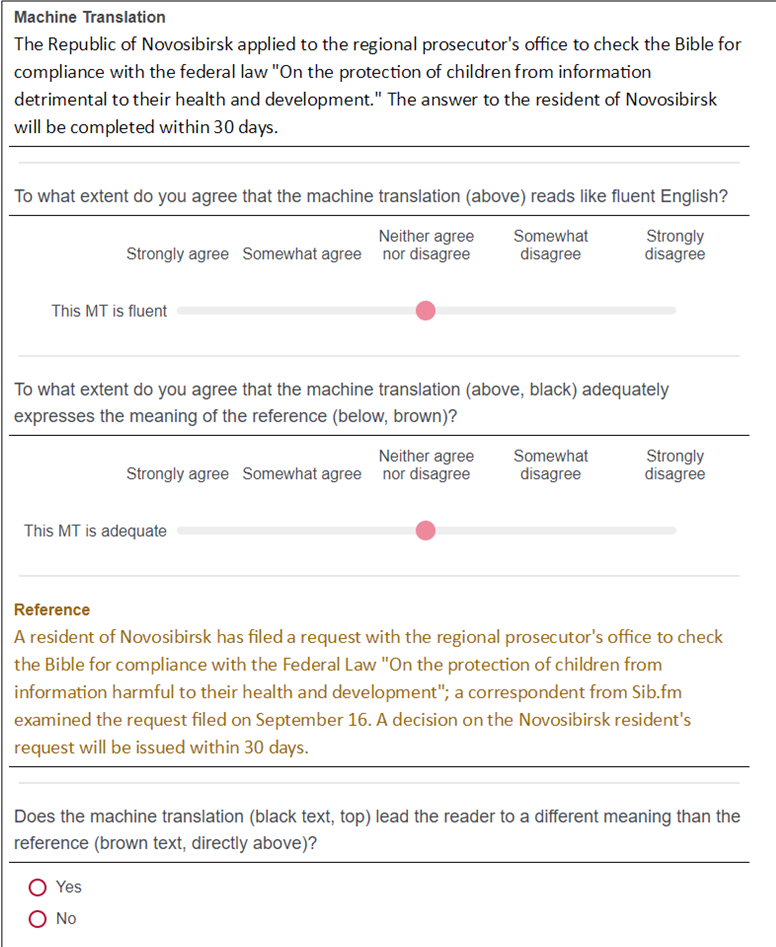}
		\caption{Example translation annotation task question}
		\label{fig3}
    \end{minipage}
\end{figure}

There were a total of 89 qualifying responses, of which 32 saw the disfluent translation, 29 saw the good translation, and 28 saw the misleading translation. The background information about the participants is summarized in Tables 1-3. In all three groups, the majority of participants were unfamiliar with human translation and a plurality of participants were unfamiliar with machine translation. This unfamiliarity with both human translation and machine translation fits our expectations for typical users of MT for assimilation. Initial perceptions of MT were primarily neutral or positive.

All of the translation examples were taken from the system outputs for the news translation task at WMT16, the 2016 Conference on Machine Translation \citep{bojar_findings_2016}. The evaluation included a ranking evaluation by MT researchers as well as a direct assessment of fluency and adequacy by crowd-workers on Amazon Mechanical Turk. Based on the direct assessment, we selected segments in three categories: ``good'' segments (high fluency, high adequacy), low fluency segments, and potentially misleading (high fluency, low adequacy) segments. 

To create a more realistic setting, one or two segments from the text surrounding each of the chosen segments in the machine translation were added. This context could provide information that clarifies (or contradicts) a single segment translation, so an annotation task was launched to verify that these longer translations matched the original labels of good, disfluent, or misleading. Figure \ref{fig3} shows an example of one of these annotation questions. Based on those annotations, we selected translations that were rated high fluency and high adequacy by all annotators as good examples. The disfluent example used in the survey was chosen based on low fluency ratings. It also had fairly low adequacy as low fluency translations were generally also rated as low adequacy, but the adequacy score was higher than fluency. The misleading example was moderately high fluency, low adequacy, and marked by all annotators as leading the reader to a different meaning than the human reference translation.

\begin{figure}[h]
\begin{subfigure}{.5\linewidth}
\centering
\includegraphics[width=.9\linewidth]{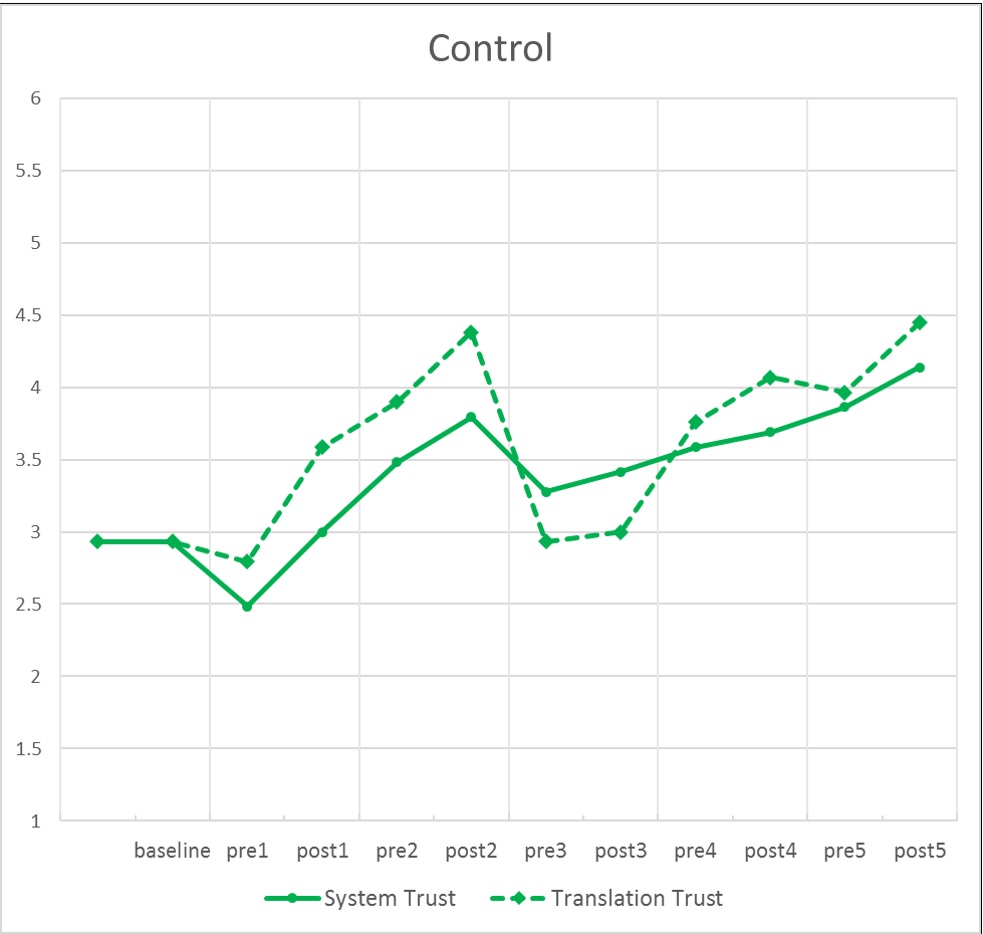}
\caption{}
\label{fig:sub1}
\end{subfigure}%
\begin{subfigure}{.5\linewidth}
\centering
\includegraphics[width=.9\linewidth]{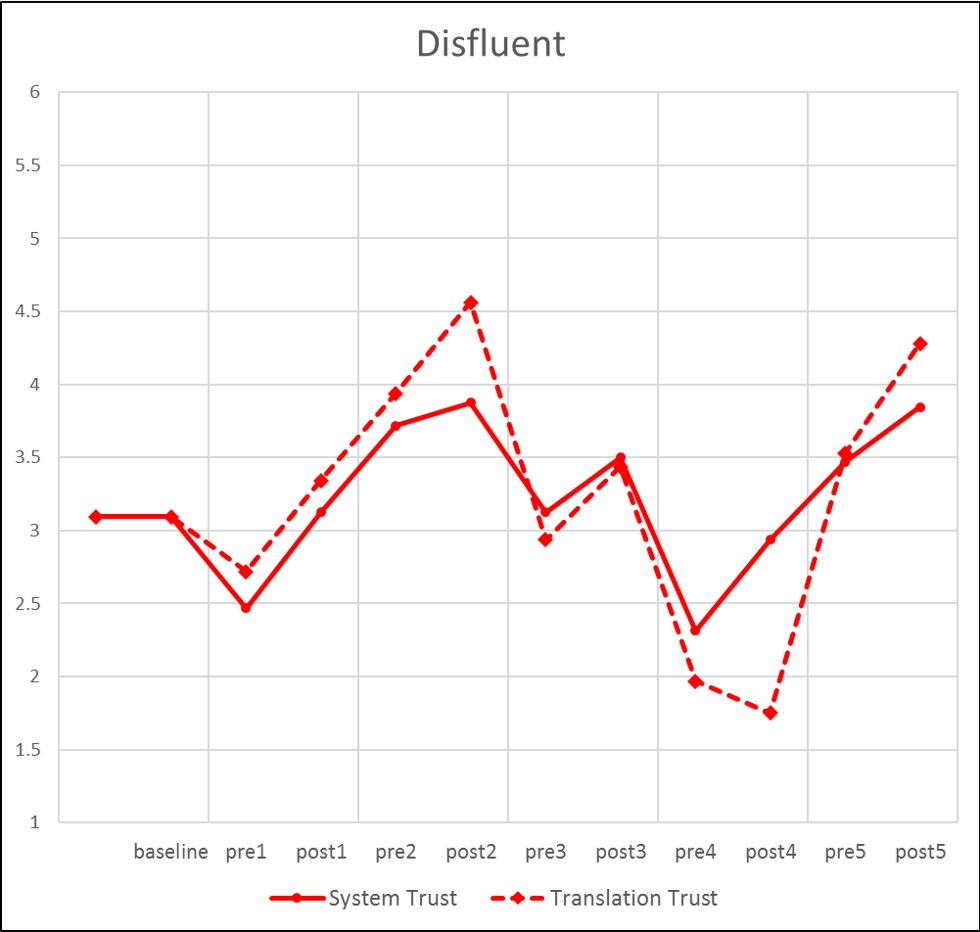}
\caption{}
\label{fig:sub2}
\end{subfigure}\\[1ex]
\begin{subfigure}{\linewidth}
\centering
\includegraphics[width=.45\linewidth]{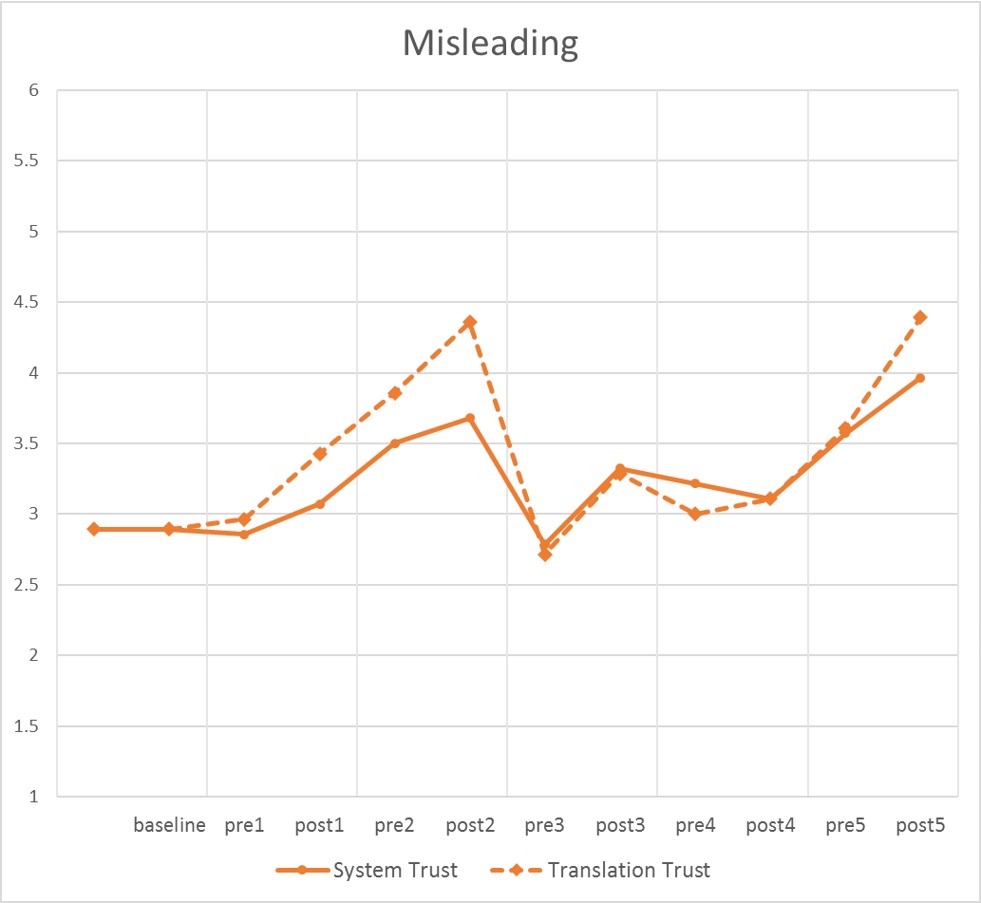}
\caption{}
\label{fig:sub3}
\end{subfigure}
\caption{System and translation trust for the control group (a), disfluent group (b), and misleading group (c)}
\label{fig4}
\end{figure}

\section{Results}

The results from the three conditions, control, disfluent, and misleading, are shown in Figure \ref{fig4} (a), (b), and (c), respectively. For each translation, the system trust score is the average of the trust scores across all responses for that setting. The “pre” score is before seeing the human translation and the “post” score is after. The “pre” translation score is the value reported by the user and the “post” is that score adjusted based on the reported change in trust.

Figure \ref{fig:sub1} shows the average translation and system trust scores for the control group. The translation trust is, not surprisingly, more volatile than the system trust. In all three conditions it basically tracks the system trust but with higher peaks and lower valleys. There are no instances where the difference between the translation trust before and after seeing the reference translation was significant ($p>0.1$). This is somewhat surprising in the case of the misleading translation which would have been expected to yield lower trust if the users understood the reference translation. Instead, the users' opinion of the misleading translation ($M=3.00$,$SD=1.76$) appears to have increased slightly when the reference translation was revealed ($M=3.11$,$SD=2.51$), however, this difference is likely statistical error as the change is not a significant increase; $t(54)=-0.18$, $p=0.85$.

Turning to system trust, with the exception of translation 3, H1 appears to hold: the trust in the system increases as the user sees more translations that are both fluent and adequate. Although translation 3 was judged as both fluent and adequate in the validation task, it appears that participants may have had difficulty understanding the translation. This confusion was not unique to those in the control group but also appeared in the other two conditions. If we set aside translation 3, the consistent increases in trust lead us to accept H1.

Figure \ref{fig:sub2} shows the results for the disfluent group. There was a significant negative change between the trust in the system after the third translation ($M=3.50$,$SD=1.27$) and after the disfluent translation ($M=2.31$,$SD=1.47$); $t(62)=3.46$, $p < 0.001$, so we can reject the null hypothesis and accept H2 for the case of fluency errors. It is interesting to note that when the final translation (fluent and adequate) was displayed the trust quickly bounced back to as high as it had been before. Figure \ref{fig:sub3}, the misleading group, also has the drop in trust on the third translation, but the misleading translation had surprisingly little effect on the users' trust. 

The change in system trust for the misleading translation ($M=-0.21$,$SD=1.50$) did not have a significant effect on user trust compared with the control; $t(55)=1.36, p=0.18$, so we cannot confirm H3: misleading translations do not appear to have a stronger effect than fluency errors. This indicates that either fluency more strongly affects user trust than adequacy does or users have a more difficult time judging adequacy than fluency. Previous research \citep{callison-burch_meta-_2007, graham_can_2017} has shown that fluency and adequacy scores are highly correlated, leading to only adequacy being measured in WMT17 \citep{bojar_findings_2017}. The correlation was taken to mean that adequacy was sufficient to judge translation quality. The results of this experiment suggest that it may in fact be the case that fluency has an out-sized effect that overshadows adequacy. The single score for adequacy should perhaps be treated as an overall quality score rather than a true adequacy score. 

\section{Limitations and Future Work} \label{limits}

This study shows that user surveys can reveal impact of different MT errors on trust. However, it is a small scale pilot study, and further work will be needed to refine the survey design, and further compare the impact of fluency and adequacy on user trust. We highlight limitations identified in this study and how they could be addressed in future work.

\paragraph{Additional Examples} A five translation sequence may not be enough to get a good sense of how trust changes over time. For instance, based on the control group, it appears that the trust levels off after a few good translations, but without more translations it is hard to know for sure. In addition, participants were exposed to a single translation example for each error category. In future experiments, having more than one series of translations would make it possible to observe whether participant reactions are consistent across distinct representative examples.

\paragraph{Double-Checking Comprehension} Asking comprehension questions before displaying the reference ensured that participants understood the translation, but there was no measure of whether they understood the reference translation. This means that those who were misled by the misleading example might not have realized that they were misled. As suggested above, having more than one possible misleading example would help, but it would also be useful to repeat the comprehension question after the reference is displayed to see whether their answers changed.

\paragraph{Include Obvious Errors} In addition to the kind of misleading example that was tested in this study, another type of error that could provide insight into how users perceive adequacy would be translations that are fluent but not adequate, but the inadequacy is obvious without even seeing the reference. Perhaps if they recognize the error in meaning immediately it will have a stronger effect on trust than if they need to read the human translation to determine adequacy.

\paragraph{Users' Involvement in Outcome} One key difference between this study and previous work on trust in automation is that this study did not provide an external motivation such as a monetary incentive (as in \citet{de_vries_effects_2003}) or an evaluation of the user (as in \citet{yang_how_2016}). It would be beneficial to conduct any future studies in an environment where the participants felt more of an investment, such as through a rewards system. The users would be more intrinsically motivated if the test were conducted in a scenario closer to the real-world experience of the participants. One example would be setting up conversations as in \citet{hara_effect_2015}. A social media scenario could be an alternative to two-way communication: participants could be given a message to promote and asked whether they would be willing to use the output of MT to post in their name in another language. Another option could be recruiting actual MT users and mimicking their work environment. This scenario would have highly motivated and informed participants but would be more difficult to create. 

\section{Conclusion}

We have argued that trust is an important metric for machine translation and proposed a survey to determine how fluency and adequacy errors affect user trust in machine translation. Because it is impossible for MT researchers and developers to educate every possible user about the strengths and weaknesses of MT, it is incumbent on the MT community to understand how users modulate their trust and use that information to develop systems that encourage an appropriate level of trust in the output.

In our study, users reacted more strongly to fluency errors than adequacy errors. If this finding holds in large-scale studies with improvements to the experimental design as outlined in section \ref{limits}, it would have implications for the development of MT systems. Although misleading translations are rare, the implicit trust users place in fluent translations obligates MT system developers to pay particular attention to adequacy. Care should be taken to reduce the frequency of misleading translations and/or to flag them as suspect and notify the user.

\section*{Acknowledgments}

We thank Jen Golbeck and Hernisa Kacorri for their feedback on this work. This work was partially supported by the Clare Boothe Luce Program.

\bibliography{trustinmt}
\bibliographystyle{acl_natbib}

\end{document}